%% file: main.tex
\documentclass[acmtog,screen,nonacm]{acmart}

\usepackage{booktabs} 

\usepackage[ruled]{algorithm2e} 

\SetAlFnt{\small}
\SetAlCapFnt{\small}
\SetAlCapNameFnt{\small}
\SetAlCapHSkip{0pt}

\input{math_definition.tex}
\newcommand{\ourmethod}{{{FabricDiffusion}}\xspace}

\usepackage{enumitem}
\usepackage{color}

\begin{document}

\title{FabricDiffusion: High-Fidelity Texture Transfer for 3D Garments Generation from In-The-Wild Clothing Images}

\author{Cheng Zhang}
\authornote{Equal contribution. {Website: \url{https://humansensinglab.github.io/fabric-diffusion.}}}
\orcid{0009-0003-6255-7648}
\affiliation{%
 \institution{Carnegie Mellon University}
 \country{United States of America}
}
\email{chzhang@tamu.edu}

\author{Yuanhao Wang}
\authornotemark[1]
\orcid{0009-0002-2737-4689}
\affiliation{%
 \institution{Carnegie Mellon University}
 \country{United States of America}
}
\email{yuanhao4@andrew.cmu.edu}

\author{Francisco Vicente Carrasco}
\orcid{0009-0005-3680-3029}
\affiliation{%
 \institution{Carnegie Mellon University}
 \country{United States of America}
}
\email{fvicente@andrew.cmu.edu}

\author{Chenglei Wu}
\orcid{0000-0002-7307-9480}
\affiliation{%
 \institution{Google Inc.}
 \country{United States of America}
}
\email{chengleiwu@gmail.com}

\author{Jinlong Yang}
\orcid{0009-0003-7605-8209}
\affiliation{%
 \institution{Google Inc.}
 \country{Switzerland}
}
\email{jinlongy@google.com}

\author{Thabo Beeler}
\orcid{0000-0002-8077-1205}
\affiliation{%
\institution{Google Inc.}
\country{Switzerland}}
\email{thabo.beeler@gmail.com}

\author{Fernando De la Torre}
\orcid{0000-0002-7086-8572}
\affiliation{%
\institution{Carnegie Mellon University}
\country{United States of America}
}
\email{ftorre@cs.cmu.edu}

\renewcommand\shortauthors{Zhang, C. et al}

\input{section/abstract}

\ccsdesc[500]{Computing methodologies}
\ccsdesc[500]{Computing methodologies~Appearance and texture representations}

\keywords{Texture transfer, BRDF material, diffusion model, synthetic data, 3D garments reconstruction}

\maketitle

\input{section/introduction}

\input{section/related}

\input{section/approach}
\input{section/exp}
\input{section/disc}

\bibliographystyle{ACM-Reference-Format}
\bibliography{main}

\appendix
\renewcommand{\appendixname}{Supplementary Material~\Alph{section}}
\setcounter{footnote}{0}
\setcounter{table}{0}
\setcounter{figure}{0}
\setcounter{section}{0}
\renewcommand\thesection{\Alph{section}}
\renewcommand{\thetable}{S\arabic{table}}  
\renewcommand{\thefigure}{S\arabic{figure}}

\vspace{5mm}
\centerline{\textbf{\LARGE{ {Supplementary Material}}}}
\vspace{5mm}

We provide details and results omitted in the main text.
\begin{itemize}
    \item Section~\ref{sec:appendixA}: Key advantages of \ourmethod.
    \item Section~\ref{sec:appendixB}: Additional details on dataset construction.
    \item Section~\ref{sec:appendixC}: Additional implementation details.
    \item Section~\ref{s_add_exp}: Additional results and analyses.
\end{itemize}

\input{section/app}

\end{document}

%% file: math_definition.tex
\usepackage{amsmath,amsfonts,bm}

\def\1{\bm{1}}

\DeclareMathAlphabet{\mathsfit}{\encodingdefault}{\sfdefault}{m}{sl}
\SetMathAlphabet{\mathsfit}{bold}{\encodingdefault}{\sfdefault}{bx}{n}

\newcommand{\R}{\mathbb{R}}

%% file: section/abstract.tex
\begin{teaserfigure}
    \centerline{\includegraphics[width=1\linewidth]{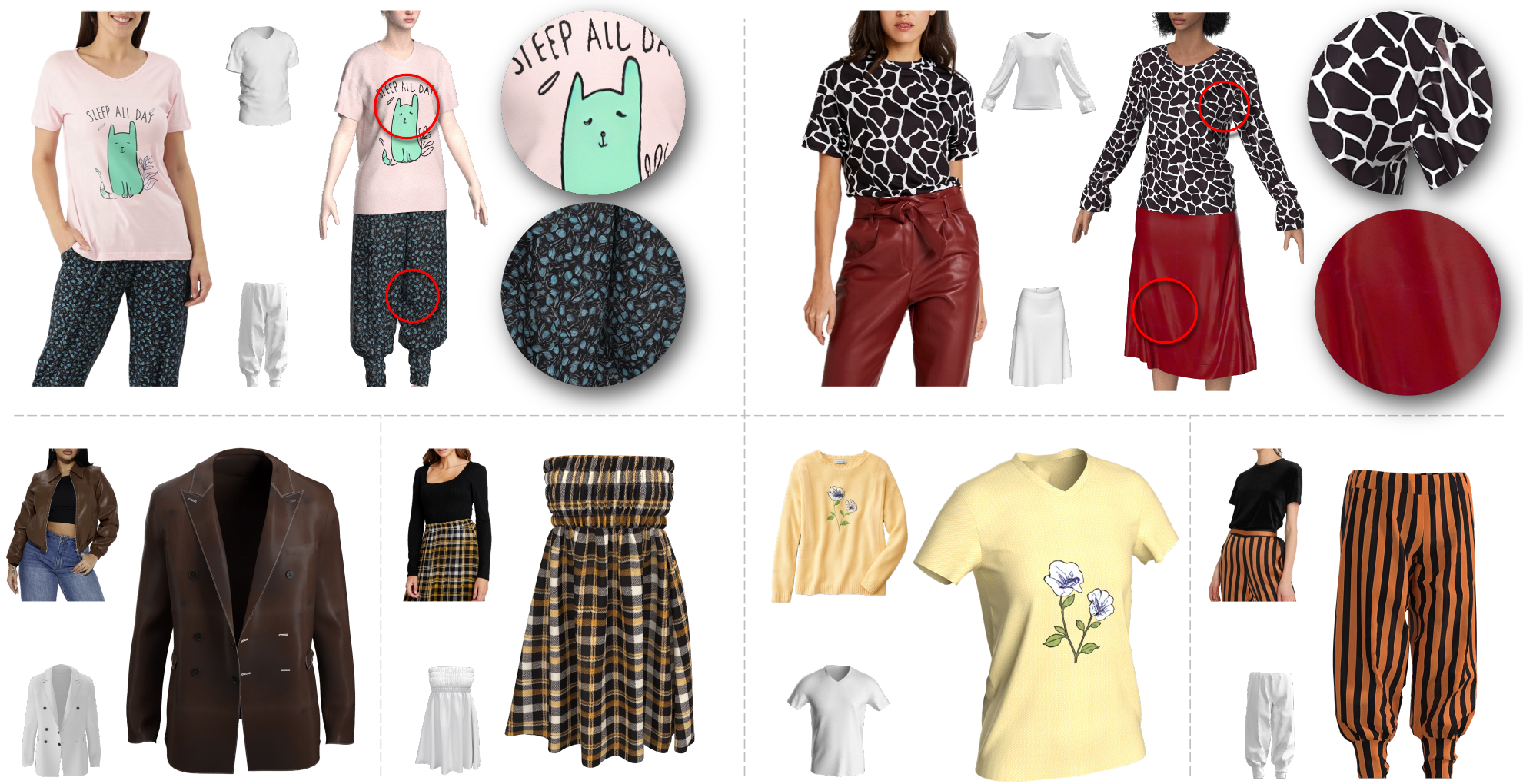}}
    \vspace{-2mm}
    \caption{\small Given a real-world 2D clothing image and a raw 3D garment mesh, we propose \ourmethod to automatically extract high-quality texture maps and prints from the reference image and transfer them to the target 3D garment surface. Our method can handle different types textures, patterns, and materials. Moreover, \ourmethod is capable of generating not only diffuse albedo but also roughness, normal, and metallic texture maps, allowing for accurate relighting and rendering of the produced 3D garment across various lighting conditions.
    }
    \label{fig:teaser} 
    \vspace{4mm}
\end{teaserfigure}

\begin{abstract}
We introduce FabricDiffusion, a method for transferring fabric textures from a single clothing image to 3D garments of arbitrary shapes. Existing approaches typically synthesize textures on the garment surface through 2D-to-3D texture mapping or depth-aware inpainting via generative models. Unfortunately, these methods often struggle to capture and preserve texture details, particularly due to challenging occlusions, distortions, or poses in the input image. Inspired by the observation that in the fashion industry, most garments are constructed by stitching sewing patterns with flat, repeatable textures, we cast the task of clothing texture transfer as extracting distortion-free, tileable texture materials that are subsequently mapped onto the UV space of the garment. Building upon this insight, we train a denoising diffusion model with a large-scale synthetic dataset to rectify distortions in the input texture image. This process yields a flat texture map that enables a tight coupling with existing Physically-Based Rendering (PBR) material generation pipelines, allowing for realistic relighting of the garment under various lighting conditions. We show that FabricDiffusion can transfer various features from a single clothing image including texture patterns, material properties, and detailed prints and logos. Extensive experiments demonstrate that our model significantly outperforms state-to-the-art methods on both synthetic data and real-world, in-the-wild clothing images while generalizing to unseen textures and garment shapes. 

\end{abstract}

%% file: section/introduction.tex
\section{Introduction}
\label{s_intro}

There is an increasing interest to experience apparel in 3D for virtual try-on applications and e-commerce as well as an increasing demand for 3D clothing assets for games, virtual reality and augmented reality applications. While there is an abundance of 2D images of fashion items online, and recent generative AI algorithms democratize the creative generation of such images, the creation of high-quality 3D clothing assets remains a significant challenge. In this work we explore how to transfer the appearance of clothing items from 2D images onto 3D assets, as shown in Figure~\ref{fig:teaser}.

Extracting the fabric material and prints from such imagery is a challenging task, since the clothing items in the images exhibit strong distortion and shading variation due to wrinkling and the underlying body shape, in addition to general illumination variation and occlusions. To overcome these challenges, we propose a generative approach capable of extracting high-quality physically-based fabric materials and prints from a single input image and transfer them to 3D garment meshes of arbitrary shapes. The result may be rendered using Physically Based Rendering (PBR) to realistically reproduce the garments, for example, in a game engine under novel environment illumination and cloth deformation.

Existing methods for example-based 3D garments texturing primarily focus on {direct texture synthesis onto 3D meshes} using techniques such as 2D-to-3D texture mapping~\cite{mir2020learning,majithia2022robust,gao2023cloth2tex} or multi-view depth-aware inpainting by distilling a pre-trained 2D generative model~\cite{richardson2023texture,zeng2023paint3d,yeh2024texturedreamer}. However, these approaches often lead to irregular and low-quality textures due to the inherent inaccuracies of 2D-to-3D registration and the stochastic nature of generative processes. Moreover, they struggle to faithfully represent texture details or disentangle garment distortions, resulting in significant degradation in texture continuity and quality.

In this work, we seek to overcome these limitations by drawing inspiration from the real-world garment creation process in the fashion industry~\cite{korosteleva2021generating,liu2023towards}: most 3D garments are typically modeled from 2D sewing patterns with normalized\footnote{We define ``normalized'' as a canonical texture space devoid of geometric distortions, illumination variations, shadows, and other inconsistencies present in the real-life input images. Terms such as ``undistored'', ``distortion-free'', ``unwarped'', and ``flat'' are used interchangeably in this paper to describe the textures free from geometric distortions.\label{footnote:normalized}} and tileable texture maps. This allows us to approach the texturing process from a novel angle, where obtaining such texture maps enables more accurate and realistic garment rendering across various poses and environments. Interestingly, if we take the 3D mesh away from our task of texture transfer, there has been a long history of development in 2D exemplar-based texture map extraction and synthesis~\cite{efros1999texture,efros2023image,wei2009state,lopes2023material,tu2022clustered,hao2023diffusion,li2022scraping,diamanti2015synthesis,cazenavette2022wearable,rodriguez2023umat,yeh2022photoscene,schroder2014image,guarnera2017woven,wu2019modeling,rodriguez2019automatic}. Nevertheless, there remains a significant gap in effectively correcting the geometric distortion or calibrating the appearance (e.g., lighting) of the fabric present in the input reference images.

How can we translate a clothing image to a normalized and tileable texture map? At first glance, solving this ill-posed inverse problem is challenging, and may require developing sophisticated frameworks to model the explicit mapping. Instead, we investigate a feed-forward pathway to simulate the texture distortion and lighting conditions from its normalized form to that on a 3D garment mesh.
Then, we propose to train a denoising diffusion model~\cite{ho2020denoising,rombach2022high} using paired texture images (i.e., both the distorted and normalized) to generate normalized and tileable texture images. Such an objective makes the training procedure fairly straightforward, which we see as a key strength. As a result, generating normalized texture images becomes solving a supervised distribution mapping problem of translating distorted texture patches back to a unified normalized space.

However, acquiring such paired training data from real clothing at scale is infeasible. 
{To address this issue, we develop a large-scale synthetic dataset comprising over $100$k textile color images, $3.8$k material PBR texture maps, $7$k prints (e.g., logos), and $22$ raw 3D garment meshes.}
These PBR textures and prints are carefully applied to the raw 3D garment meshes and then rendered using PBR techniques under diverse lighting and environmental conditions, simulating real-world scenarios. For each fabric captures from the textured 3D garment, we render a corresponding image using ground-truth PBR textures, which are applied to a flat mesh under a controlled illumination condition, i.e., orthogonal close-up views with a pointed lighting from above. The captured texture inputs along with their ground-truth flat mesh render are used to train our diffusion model. Figure~\ref{fig:data_gen} illustrates the pipeline of training data construction.

We name our method \ourmethod and systematically study the performance on both synthetic data and real-world scenarios. Despite being trained entirely on synthetic rendered examples, \ourmethod achieves zero-shot generalization to in-the-wild images with complex textures and prints. Furthermore, the outputs of \ourmethod seamlessly integrate with existing PBR material estimation pipelines~\cite{sartor2023matfusion}, allowing for accurate relighting of the garment under different lighting conditions. 
In summary, \ourmethod represents a state-of-the-art approach capable of extracting undistorted texture maps from real-world clothing images to produce realistic 3D garments.

%% file: section/related.tex
\section{Related Work}\label{s_related}

Our method built upon recent and seminar work on image-based 3D garment modeling, exemplar-based texture and material extraction, and diffusion-based image generation.

\subsection{Image-based 3D Garment Modeling}
\subsubsection{Image-to-mesh texture transfer.}
Existing methods on 2D-to-3D texture transfer typically involve (1) learning a 2D-to-3D registration~\cite{mir2020learning,majithia2022robust,gao2023cloth2tex} and (2) conducting depth-aware inpainting supervised by a pre-trained image generative model~\cite{rombach2022high} to guarantee multi-view consistency~\cite{richardson2023texture,zeng2023paint3d,yeh2024texturedreamer,zhang2024mapa}. However, these methods often fail to capture the high frequency details of the texture or leads to irregular textures. In this work, we tackle the problem of texturing 3D garments from a drastically different angle, aiming to extract normalized texture maps from a single real-life clothing image so that we can easily apply them to the 2D UV space (i.e., sewing pattern~\cite{korosteleva2021generating}) of the 3D garment mesh for realistic rendering.

\subsubsection{Image-based sewing pattern generation.}
We argue that a major cause of the quality gap observed in generated textures is not the capacity of the generation networks, but rather from a suboptimal choice of representations for the texture generation operating from the reference image to the 3D mesh. Unfortunately, there has been little progress in leveraging the idea of generating texture maps that can be used in the 2D UV space, despite the availability of sewing patterns for 3D garments as the sewing pattern can either be manually created by technical artists~\cite{liu2023towards} or automatically reconstructed from reference images~\cite{liu2023towards,li2023diffavatar,chen2022structure}. Concurrently, DeepIron~\cite{kwon2023deepiron} is the only work that leverages the similar idea of transferring the texture using sewing pattern representation. Unlike our method, they aim to transfer entire garments without PBR texture maps and exhibits subpar performance in real-world scenarios for practical usages.

\subsubsection{3D garment generation.}
Recently, there has been growing interest in 3D garment generation using generative models. For instance, GarmentDreamer~\cite{li2024garmentdreamer} and WordRobe~\cite{srivastava2024wordrobe} are recent work that focus on text-based garment generation, whereas our approach transfers textures using image guidance. Another relevant work, Garment3DGen~\cite{sarafianos2024garment3dgen}, can reconstruct both textures and geometry from a single input image. However, unlike Garment3DGen, our work focuses on generating distortion-free texture and prints and has the additional capability of generating standard PBR materials.

\subsection{Exemplar-based Texture and Material Extraction}

The literature on exemplar-based texture and material extraction is vast. We focus on representative works that are related to ours.

\subsubsection{Texture map extraction.}
We recast the task of image-to-3D garment texture transfer as generating texture maps from reference clothing image patches. {\citet{hao2023diffusion} trained a diffusion model to rectify distortions and occlusions in natural texture images. However, it does not extract tileable texture patches or PBR materials for fabrics.} {More recently}, Material Palette~\cite{lopes2023material} addressed a similar problem by using a diffusion-based generative model to extract PBR materials. Their approach relies on personalization methods such as textual inversion~\cite{gal2022image} to represent the exemplar patch without normalizing the patch into a canonical space, i.e., distortion-free with unified lighting.

\subsubsection{Tileable texture synthesis.}
Previous work have attempted to synthesize tileable textures with a variety of methods, such as by maximizing perceived texture stationary \cite{moritz2017texture}, by using Guided Correspondence \cite{zhou2023neural}, by finding repeated patterns in images using pre-trained CNN features \cite{rodriguez2019automatic}, by manipulating the latent space of pre-trained GANs \cite{rodriguez2022seamlessgan}, or by modifying the noise sampling process of a diffusion model, i.e., rolled-diffusion \cite{vecchio2023controlmat}. We found that a simple circular padding strategy following ~\cite{zhou2022tilegen} performs well with our model architecture for addressing tileable texture generation.

\subsubsection{BRDF material estimation.}
A significant body of research exists on BRDF material estimation from a single image \cite{deschaintre2018single, henzler2021generative, vecchio2021surfacenet, casas2023smplitex, vecchio2024matsynth, vecchio2024matfuse}.
Our model produces normalized texture maps in a canonical space, enabling compatibility with existing Bidirectional Reflective Distribution Function (BRDF) material estimation pipelines such as MatFusion~\cite{sartor2023matfusion}, which can be integrated seamlessly with our output normalized textures. By fine-tuning the pre-trained MatFusion model with fabric PBR texture data and incorporate it into our pipeline, our model generates high-quality material maps for realistic 3D garment rendering.

\subsection{Diffusion-based Image Generation}
Our model architecture is inspired by the recent advancements in diffusion-based image generation models~\cite{ho2020denoising,rombach2022high,sohl2015deep}. In this work, we fine-tune the pre-trained image generative model using carefully created synthetic data, enabling texture normalization, which includes distortion removal, lighting calibration, and shadow elimination.

%% file: section/approach.tex
\section{Method}

\begin{figure*}[t]
     \centerline{\includegraphics[width=0.98\linewidth]{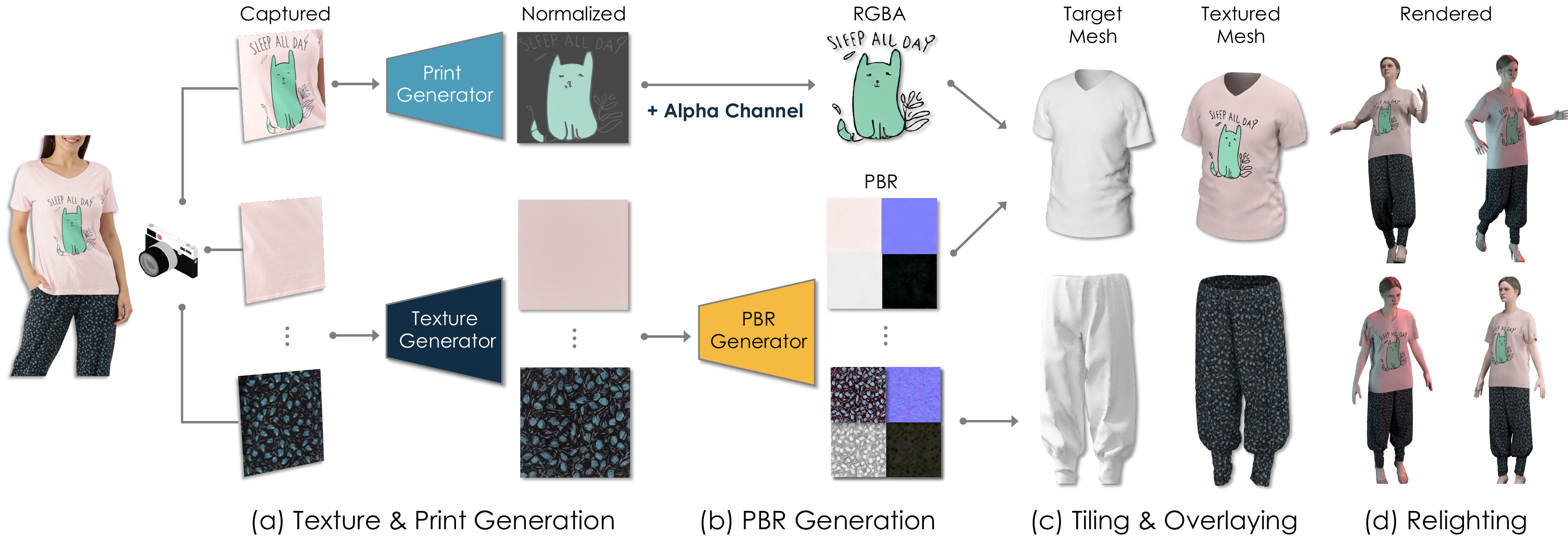}}
    \caption{\small Overview of FabricDiffusion. Given a real-life clothing image and region captures of its fabric materials and prints, (a) our model extracts normalized textures and prints, and (b) then generates high-quality Physically-Based Rendering (PBR) materials and transparent prints. (c) These materials and prints can be applied to the target 3D garment meshes of arbitrary shapes (d) for realistic relighting. Our model is trained purely with synthetic data and achieves zero-shot generalization to real-world images.
    }
    \label{fig:overview} 
\end{figure*}

We propose \ourmethod to extract normalized, tileable texture images and materials from a real-world clothing image, and then apply them to the target 3D garment. The overall framework is illustrated in Figure~\ref{fig:overview}. We first introduce the problem statement in Section~\ref{s_method_statement}, followed by procedures for constructing synthetic training examples in Section~\ref{s_method_data}. In Section~\ref{s_method_fabricdiffusion}, we detail our specific approach of texture map generation. Finally, we describe PBR materials generation and garment rendering in Section~\ref{s_method_pbr}.

\subsection{Problem Statement}\label{s_method_statement}
Given an input clothing image $I$ and a captured texture region $x$, which may exhibit various distortions and illuminations due to occlusion and poses present in the input image, our goal is learn a mapping function $g$ that takes the captured patch $x$ and outputs the corresponding normalized texture map $\Tilde{x}$, effectively correcting the distortions. The texture map $\Tilde{x}$ needs to retain the intrinsic properties of the original captured region, such as color, texture pattern, and material characteristics.

As mentioned in Section~\ref{s_intro}, we formulate the generation of normalized texture maps from a real-life clothing patch as a distribution mapping problem. Specifically, the mapping function $g$ can be modeled by a generative process:
\begin{equation}\label{eq:statement}
    \Tilde{x} \sim  G_{\theta}(x, \epsilon), \epsilon \sim \mathcal{N} (0, \mathbf{I}),
\end{equation}
where the generative model $G_{\theta}$, parameterized by $\theta$, takes the input patch $x$ as a condition and samples from Gaussian noise to generate the distortion-free texture map $\Tilde{x}$ in a canonical space. To train the generator $G$, we must create a large number of paired training examples $(x, x_0)$ across various types of textures. Here $x$ is the input capture and $x_o$ is the corresponding ground-truth normalized texture. After the model training, we expect to align the sampled output $\Tilde{x}$ with the distribution of normalized textures.

\subsection{Synthetic Paired Training Data Construction}\label{s_method_data}
Collecting paired training examples with real clothing poses significant challenges. 
In contrast, we found that PBR textures --- the fundamental unit for appearance modeling in 3D apparel creation --- are much more accessible from public sources (see Section~\ref{s_exp_setup} for details on dataset collection). 
Given these observations, we propose to build synthetic environments for constructing distorted and flat rendered training pairs using the PBR material model~\cite{mcauley2012practical}. Figure~\ref{fig:data_gen} illustrates the overall pipeline.

\subsubsection{Paired training examples construction.}
For each material, we collect the ground-truth diffuse albedo ($k_d \in \R^3$), normal ($k_n \in \R^3$), roughness ($k_r \in \R^2$), and metallic ($k_m \in \R^2$) material maps. To create distorted rendered images that mimic real-world surface deformation and lighting, we map these material maps onto a raw garment mesh sampled from 22 common garment types. The PBR textures are tiled appropriately and illuminated using four environment maps with white lights to avoid color biases. During rendering, we capture frontal views of the garment and randomly crop patches from the rendered images to match the original fabric texture size.

\begin{figure}[t]
    \centerline{\includegraphics[width=0.98\linewidth]{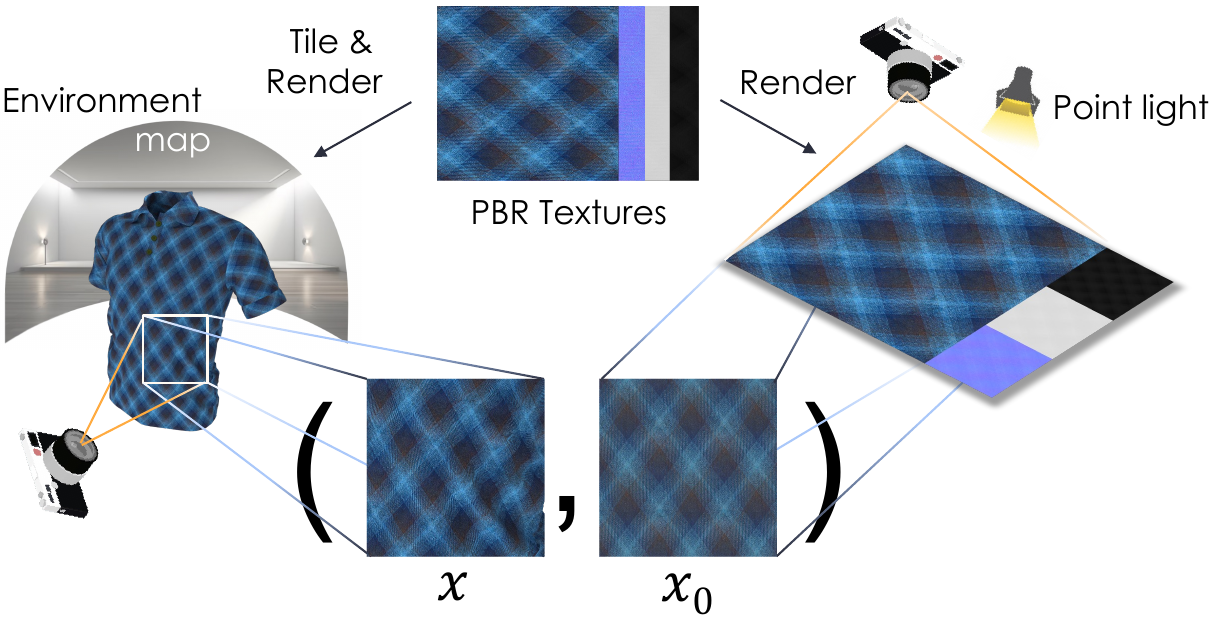}}
    \caption{\small Pipeline of paired training data construction. Given the textures of a PBR material, we apply them to both the target raw 3D garment mesh and the plain mesh. The 3D garment is rendered using an environment map, while the plain mesh is illuminated using a point light from above. The resulting rendered images $(x, x_0)$ from both meshes serve as the paired training examples for training our texture generative model (Section~\ref{s_method_data}).
    }
    \label{fig:data_gen} 
\end{figure}

Separately, we render the same texture material on a plane mesh to create flat rendered images as ground-truths (image $x_0$ in Figure~\ref{fig:data_gen}). For illumination, we use a fixed point light above the surface center and a fixed orthogonal camera for rendering. This method is highly beneficial as it provides supervision to align the distorted rendered images on the 3D garment to a canonical space of normalized, flat images with a unified lighting condition.

In fact, our flat image rendering and capturing approach may be reminiscent of the input format used in well-known SVBRDF material estimation methods~\cite{sartor2023matfusion,zhou2021adversarial,zhou2022tilegen,zhou2023photomat}, which require orthogonal close-up views of the materials and/or a flashing image as input. As will be described in Section~\ref{s_method_pbr}, the output normalized textures from our method can be effectively integrated with SVBRDF material estimation models to generate high-quality PBR material maps. 

\subsubsection{Paired prints (e.g., logos) construction.}
In additional to general textures, we aim to transfer clothing details by creating warped and flat pairs of print images. We map the print to a random location on the garment mesh and blend it with a uniformly colored background texture. Unlike flat texture generation on a plane mesh, we use the original print image with a transparent background as the flat image.

\subsubsection{Scaling up training data with Pseudo-BRDF materials.}
While the texture material maps are easier to acquire than real clothing, we raise the question: Do we really need a large amount of real BRDF material maps for paired training data construction, and what if we cannot obtain enough data?

In this work, we are able to collect a BRDF dataset comprises 3.8k assets in total (see Section~\ref{s_exp_setup} for details), covering a broad spectrum of fabric materials. However, the texture patterns in this dataset exhibit limited diversity because it is not large enough to model the appearance of fabric textures in our real life, given the vast range of colors, patterns, and materials. To address this, we augmented the dataset by gathering 100k textile color images featuring a wide array of patterns and designs, which are then used to generate pseudo-BRDF\footnote{{Since the normal, roughness, and metallic maps of the 100k textile images are sampled instead of ground truth, they are referred to as pseudo-BRDF data.}} materials. Specifically, the color image served as the albedo map, while the roughness map was assigned a uniform value $\alpha$ sampled from the distribution $\mathcal{N}(0.708, 0.193^2)$, with 0.708 and 0.193 representing the population mean and standard deviation of the mean roughness values of the real BRDF dataset, respectively. The metallic map was assigned a uniform value $\max(\beta, 0)$, where $\beta \sim \mathcal{U}(-0.05,0.05)$, and the normal map was kept flat.

We use a combination of real (3.8k) and pseudo-BRDF (100k) materials to create paired rendered images for training our texture generation model. During paired training examples construction, both real and pseudo-BRDF have \( x \) and \( x_0 \) (as illustrated in Figure~\ref{fig:data_gen}), representing distorted and flat textures, respectively. Intuitively, the primary goal of our texture generator is to eliminate geometric distortions, and our generated {pseudo} rendered images, serve this purpose effectively.

\subsection{Normalized Texture Generation via \ourmethod}
\label{s_method_fabricdiffusion}
Given the paired training images, we build a denoising diffusion model to learn the distribution mapping from the input capture to the normalized texture map. Next, we detail our training objective, model architecture and training, and the design for tileable texture generation and alpha-channel-enabled\footnote{Alpha-channel-enabled prints are images with transparency that can be overlaid onto existing images for realistic composition and rendering.} prints generation.

\subsubsection{Training objective of conditional diffusion model.}
Diffusion models~\cite{sohl2015deep,ho2020denoising} are trained to capture the distribution of training images through a sequential Markov chains of adding random noise into clean images and denoising pure noise to clean images. We leverage Latent Diffusion Model (LDM)~\cite{rombach2022high} to improve the efficiency and quality of diffusion models by operating in the latent space of a pre-trained variational autoencoder~\cite{kingma2013auto} with encoder $\mathcal{E}$ and decoder $\mathcal{D}$. In our case, given the paired training data $(x, x_0)$, where $x$ is the distorted patch and $x_0$ is the normalized texture, the feed-forward process is formulated by adding random Gaussian noise into the latent space of image $x_0$:
\begin{equation}
x_t = \sqrt{\gamma(t)} \mathcal{E}(x_0) + \sqrt{1-\gamma(t)} \epsilon,
\label{eq:add_noise}
\end{equation}
where $x_t$ is a noisy latent of the original clean input $x_0$, $\epsilon \sim \mathcal{N}(0, \mathbf{I})$, $t \in [0,1]$, and $\gamma(t)$ is defined as a noise scheduler that monotonically descends from 1 to 0.
By adding the distorted image $x$ as the condition, the reverse process aims to denoise Gaussian noises back to clean images by iteratively predicting the added noises at each reverse step. We minimize the following latent diffusion objective:
\begin{equation}
L(\theta) = \mathbb{E}_{\mathcal{E} (x), \epsilon \sim \mathcal{N}(0, \mathbf{I}), t} \left[ \left\| \epsilon - \epsilon_{\theta}({x}_t, t, \mathcal{E}(x)) \right\|^2 \right],
\label{eq:diffusion_loss}
\end{equation}
where $\epsilon_{\theta}$ denotes model parameterized by a neural network, $x_t$ is the noisy latent for each timestep $t$, and $\mathcal{E}(x)$ is the condition. 

Recalling Equation~\ref{eq:statement}, the above formulation incorporates input-specific information (i.e., the captured patch $x$) into the training process for generating normalized textures. As will be shown in the experimental results in Section~\ref{s_exp_main}, this design is the key to producing faithful texture maps that differs from existing per-example optimization-based texture extraction approaches~\cite{lopes2023material,richardson2023texture}.

\subsubsection{Model architecture and training.}
Any diffusion-based architecture for conditional image generation can realize Equation~\ref{eq:diffusion_loss}. Specifically, we use Stable Diffusion~\cite{rombach2022high}, a popular open-source text-conditioned image generative model pre-trained on large-scale text and image pairs. To support image conditioning, we use additional input channels to the first convolutional layer, where the latent noise $x_t$ is concatenated with the conditioned image latent $\mathcal{E}(x)$. The model's initial weights come from the pre-trained Stable Diffusion v1.5, while the newly added channels are initialized to zero, speeding up training and convergence. We eliminate text conditioning, focusing solely on using a single image as the prompt. This approach addresses the challenge of generating normalized texture maps, which text prompts struggle to describe accurately~\cite{deschaintre2023visual}.

\subsubsection{Circular padding for seamless texture generation.}
To ensure the generated texture maps are tileable, we employ a simple yet effective circular padding strategy inspired by TileGen~\cite{zhou2022tilegen}. Unlike TileGen, which uses a StyleGAN-like architecture~\cite{karras2020analyzing} and needs to replace both regular and transposed (e.g., upsampling or downsampling) convolutions, we only apply circular padding to all regular convolutional layers, thanks to the flexibility of diffusion models.

\subsubsection{Transparent prints generation.}
The vanilla Stable Diffusion model can only output RGB images, lacking the capability to generate layered or transparent images, which is in stark contrast to our demand for prints transfer. Instead of redesigning the existing generative model~\cite{zhang2024transparent}, we propose a simple and effective recipe to post-process the generated RGB print images for computing an additional alpha channel. We hypothesize that the alpha map for prints can be approximated as binary -- either fully transparent or fully opaque. Based on this assumption, we assign a new RGB value for each pixel $(i, j)$ as follows: 
\begin{equation}
    \text{RGB} (i, j) = \max \Bigl[ 0, \frac{\Tilde{x}(i, j) - 0.1}{0.9} \Bigr],
\end{equation}
where $\Tilde{x}$ is the generated texture (Equation~\ref{eq:statement}). The alpha channel value at each pixel $(i, j)$ is thus determined by the following criteria: 
\begin{equation}
    \text{A}(i, j) = 
        \begin{cases}
            \qquad 1 &              \text{if} ~ \Tilde{x} (i, j) \geq 0.1, \\ 
            \Tilde{x}(i, j) / 0.1 & \text{otherwise}.
        \end{cases}
\end{equation}
This approach assigns full opacity (alpha value of 1) to pixels where the initial value exceeds a certain threshold, and scales down the alpha value for other pixels, designating them as transparent background. As will be shown in Section~\ref{s_exp_main} and  Figure~\ref{fig:print_real}, our method can handle complex prints and logos and output RGBA print images that can be overlaid onto the fabric texture.

\subsection{PBR Materials Generation and Garment Rendering}\label{s_method_pbr}
Our \ourmethod model is able to generate a normalized texture map that is tileable, flat, and under a unified lighting, ensuring compatibility with the SVBRDF material estimation method. The goal of this work is not to develop a new material estimation method but to demonstrate the compatibility of our approach with existing methods. MatFusion~\cite{sartor2023matfusion} is a state-of-the-art model trained on approximately 312k SVBRDF maps, most of which are non-fabric or non-clothing materials. We fine-tune this model using our dataset of real fabric BRDF materials. Specifically, we use our normalized textures as inputs, with the material maps $(k_d, k_n, k_r, k_m)$ as ground-truths for model fine-tuning. 

The generated PBR material maps can be used for tiling in the garment sewing pattern. The remaining question is {how to determine the scale for tiling?} We consider two specific strategies: (1) Proportion-aware tiling. We use image segmentation to calculate the proportion of the caputured region relative to the segmented clothing, maintaining a similar ratio when tiling the generated texture onto the sewing pattern. (2) User-guided tiling. We emphasize that an end-to-end automatic tilling method may not be optimal, as user involvement is often necessary to resolve ambiguities and provide flexibility in fashion industries. 

%% file: section/exp.tex
\section{Experiments}
We validate \ourmethod with both synthetic data and real-world images across various scenarios. We begin by introducing the experimental setup in Section~\ref{s_exp_setup}, followed by detailing the experimental results in Section~\ref{s_exp_main}. Finally, we conduct ablation studies and show several real-world applications in Section~\ref{s_exp_ablation}.

\subsection{Setup}
\label{s_exp_setup}

\subsubsection{Dataset.}
We detail the process of collecting BRDF texture, print, and garment datasets. 
(1) Fabric BRDF dataset. This dataset includes 3.8k real fabric materials and 100k pseudo-BRDF textures (RGB only). We reserved 200 real BRDF materials for testing the PBR generator and 800 pseudo-BRDF materials (combined with the 200 real materials) for testing the texture generator.
(2) 3D garment dataset. We collected 22 3D garment meshes for training and 5 for testing. Using the method in Section~\ref{s_method_data}, we created 220k flat and distorted rendered image pairs for training and 5k pairs for testing.
(3) Logos and prints dataset. This dataset contains 7k prints and logos in PNG format. We generated pseudo-BRDF materials with specific roughness and metallic values and a flat normal map. Dark prints were converted to white if necessary. By compositing these onto 3D garments, we produced 82k warped print images.

\subsubsection{Evaluation protocols and tasks.}
We compare \ourmethod to state-of-the-art methods on two tasks: 
(1) Image-to-garment texture transfer. Our ultimate goal is to transfer the textures and prints from the reference image to the target garment. We evaluate \ourmethod and compare it to baseline methods using both synthetic and real-world test examples.
(2) PBR materials extraction. We provide both quantitative and qualitative results on PBR materials estimation using our testing BRDF materials dataset.

\subsubsection{Evaluation metrics}
We evaluate the quality of generated textures and garments using commonly used metrics: LPIPS~\cite{zhang2018unreasonable} , SSIM~\cite{wang2004image}, MS-SSIM~\cite{wang2003multiscale}, DISTS~\cite{ding2020image}, and FLIP~\cite{andersson2020flip}. To evaluate the tileability of the generated textures, we adopt the metric proposed by TexTile
\cite{rodriguez2024textile}.
For the image-to-garment texture transfer task, we additionally report FID~\cite{heusel2017gans} and CLIP-score in CLIP image feature space~\cite{radford2021learning,gal2022image} to evaluate the visual similarity of the textured garment with the original input clothing.

\subsubsection{Baseline methods.}
We compare with state-of-the-art methods that support image-to-mesh texture transfer, including: (1) TEXTure~\cite{richardson2023texture}, the most representative method for texturing a 3D mesh based on a small set of sample images through per-subject optimization (i.e., textual inversion~\cite{gal2022image} for personalization). (2) Material Palette~\cite{lopes2023material}, which focuses on texture extraction and PBR materials estimation from a single image using generative models. (3) MatFusion~\cite{sartor2023matfusion}, for PBR materials estimation for general materials, not specifically fabric or clothing. We fine-tuned the pre-trained MatFusion model with our curated fabric BRDF training examples, resulting in improved performance.

\begin{figure*}[t]
    \centerline{\includegraphics[width=0.9\linewidth]{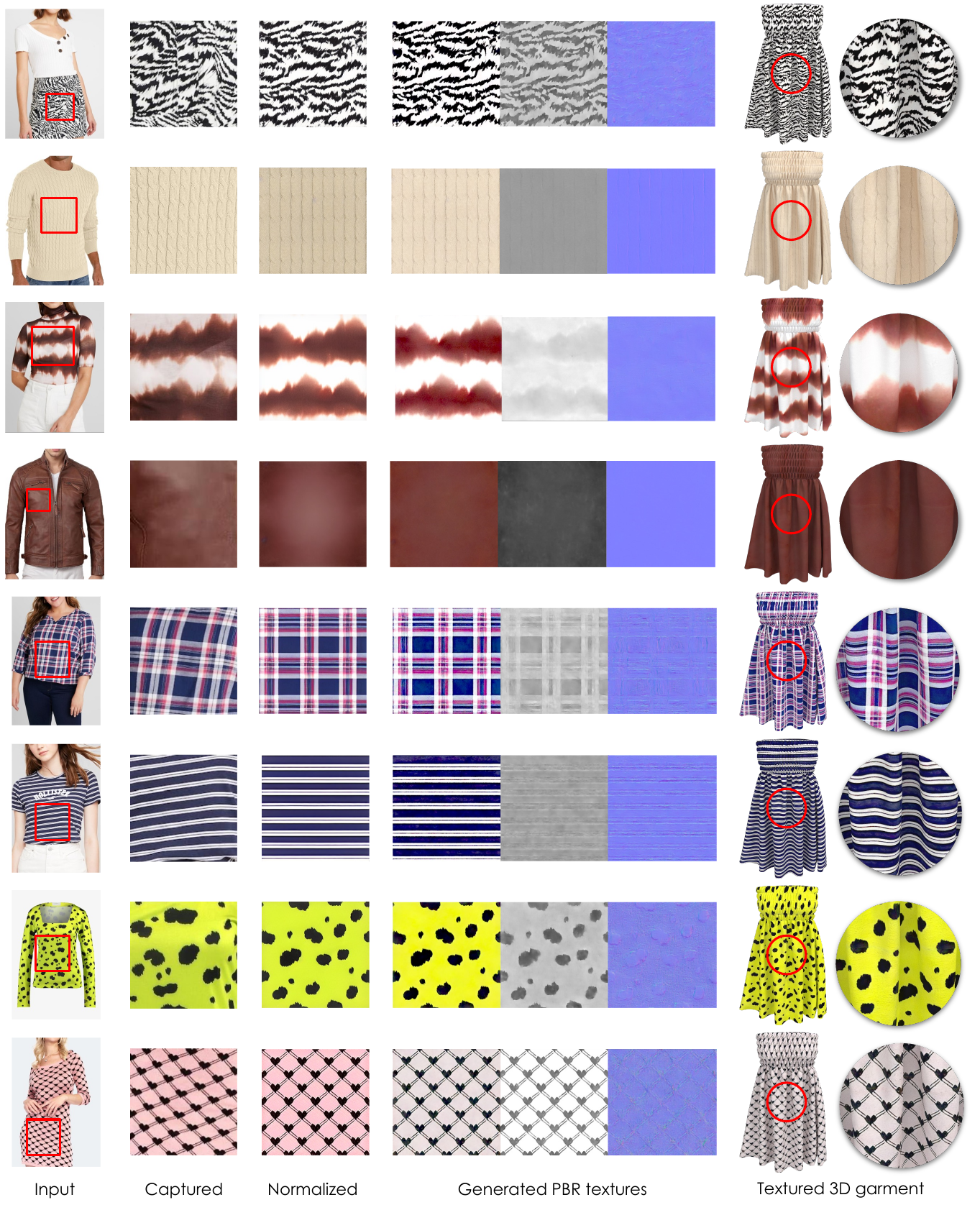}}
    \vspace{-3mm}
    \caption{\small Results on texture transfer on real-world clothing images. Our method can handle real-world garment images to generate normalized texture maps, along with the corresponding PBR materials. The PBR maps can be applied to the 3D garment for realistic relighting and rendering. 
    }
    \label{fig:exp_real} 
\end{figure*}

\begin{figure*}[t]
     \centerline{\includegraphics[width=0.98\linewidth]{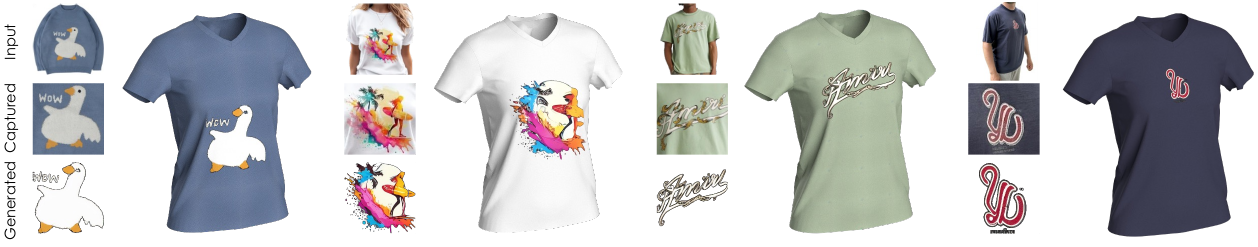}}
     \vspace{-3mm}
    \caption{\small Results on prints and logos transfer on real-world images. Given a real-life garment image with prints and/or logos, and the cropped patch of the region where the print is located. Our method generates a distortion-free and transparent print element, which can be applied to the target 3D garment for realistic rendering. Note that the background texture is transferred using our method as well.}
    \label{fig:print_real} 
\end{figure*}

\begin{figure}[t]
     \centerline{\includegraphics[width=0.98\linewidth]{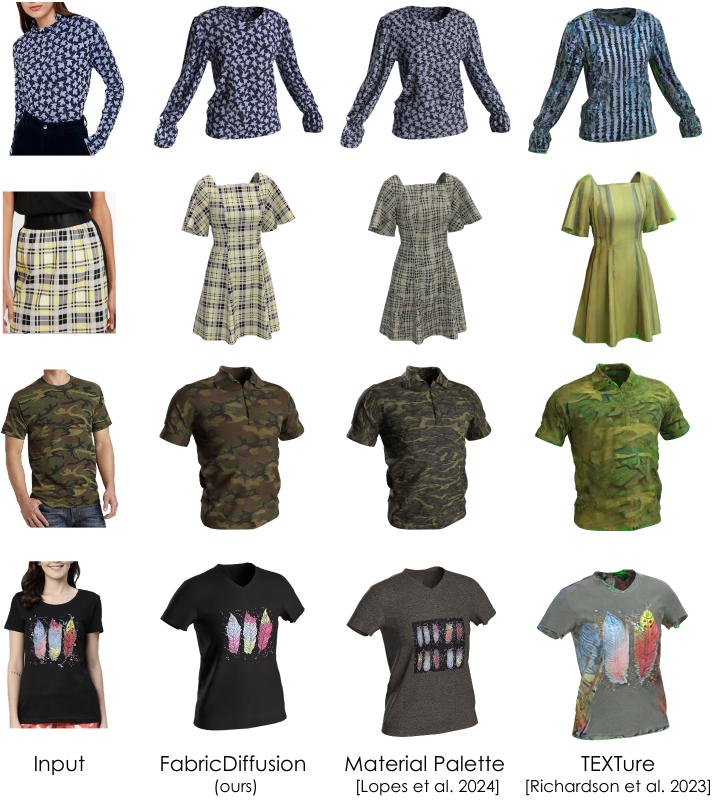}}
    \caption{\small Comparison on image-to-garment texture transfer. \ourmethod faithfully captures and preserves the texture pattern from the input clothing. We observe texture irregularities and artifacts for Material Palette~\cite{lopes2023material} and TEXTure~\cite{richardson2023texture}.
    }
    \label{fig:exp_sota} 
\end{figure}

\subsection{Experimental Results}\label{s_exp_main}
\subsubsection{\ourmethod on real-world clothing images.}
We first show the results of our method on real-world images in Figure~\ref{fig:exp_real}. Our method effectively transfers both texture patterns and material properties from various types of clothing to the target 3D garment. Notably, our method is capable of recovering challenging materials such as knit, translucent fabric, and leather. 
We attribute this success to our construction of paired training examples that seamlessly couples the PBR generator with the upstream texture generator.
Since we focus on non-metallic fabrics, the metallic map is omitted in the visualizations in the section. Please be referred to Appendix for more details and results.

\subsubsection{\ourmethod on detailed prints and logos.}
In addition to texture patterns and material properties, our \ourmethod model can transfer detailed prints and logos. Figure~\ref{fig:print_real} shows some examples. We highlight two key advantages of our design that benefit the recovery of prints and logos. First, our conditional generative model corrects geometry distortion caused by human pose or camera perspective. Second, as detailed in Section~\ref{s_method_fabricdiffusion}, our method can generate prints with a transparent background, enabling practical usage in garment appearance modeling.

\subsubsection{Image-to-garment texture transfer.}
In Figure~\ref{fig:exp_sota}, we compare our method with Material Palette~\cite{lopes2023material} and TEXTure~\cite{richardson2023texture} for image-to-garment texture transfer. We present the results on real-world clothing images featuring a variety of textures, ranging from micro to macro patterns and prints. Our observations indicate that \ourmethod not only recovers repetitive patterns, such as scattered stars or camouflage, but also maintains the regularity of structured patterns, like the plaid on a skirt. Please refer to Table~\ref{tab:sota_entire_clothing} for quantitative results.

\input{table/tab_rendered_clothing}


\input{table/tab_pbr}


\begin{figure}[t]
     \centerline{\includegraphics[width=0.98\linewidth]{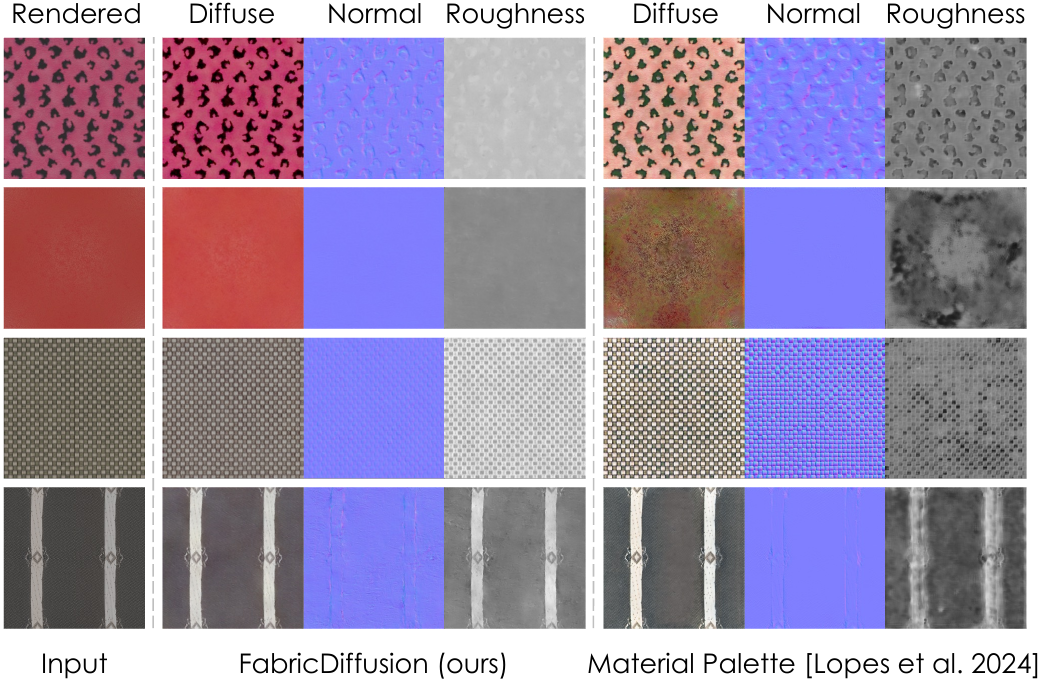}}
    \vspace{-3mm}
    \caption{\small Qualitative comparison on PBR materials extraction. Material Palette~\cite{lopes2023material} can hardly capture fabric materials while our \ourmethod model is capable of recovering physical properties for fabric textures especially on roughness and diffuse maps.
    }
    \label{fig:exp_pbr} 
\end{figure}

\subsubsection{PBR materials extraction.}
We compare our method to Material Palette~\cite{lopes2023material} and MatFusion~\cite{sartor2023matfusion} on PBR materials extraction. In Table~\ref{tab:sota_pbr}, we present a comparison of pixel-level MSE and SSIM between the generated material maps and the ground-truths. Our \ourmethod material generator, fine-tuned from the base MatFusion model with additional fabric BRDF training examples, demonstrates superior performance. Additionally, Figure~\ref{fig:exp_pbr} shows visual comparisons between \ourmethod and Material Palette. While Material Palette~\cite{lopes2023material} struggles to accurately capture fabric materials, our \ourmethod model excels in recovering the physical properties for fabric textures, particularly in roughness and diffuse maps. 
We also evaluate different methods on the rendered images and show the results in Table~\ref{tab:rendered_image}. Particularly, we use render-aware metrics like FLIP~\cite{andersson2020flip} and perceptual metrics like LPIPS and DISTS. \ourmethod consistently achieve better performance over other approaches.

\input{table/tab_render_metrics}

\subsection{Ablations, Analyses, and Applications}\label{s_exp_ablation}

\subsubsection{Ablation on circular padding and tileability analysis.}
{
We conduct an ablation study to evaluate the impact of circular padding using the TexTile metric~\cite{rodriguez2024textile}, where higher values indicate better tileability. The results show that the MaterialPalette~\cite{lopes2023material} achieves a score of 0.54. Our method without circular padding scores 0.47, while with circular padding, our method improves significantly, reaching a score of 0.62.
}

\subsubsection{Ablation on pseudo-BRDF data.}
We compare the performance of using combined real-BRDF and pseudo-BRDF data versus using only real-BRDF data. The results, summarized in Table \ref{tab:pseudo-BRDF}, demonstrate that the inclusion of pseudo-BRDF data alongside real-BRDF data improves performance across all metrics.

\subsubsection{Effect of the capture location.}
In Section~\ref{s_method_pbr}, we explored how \ourmethod can be integrated into an end-to-end framework for 3D garment design. To assess whether the generated texture remains consistent with the input, Figure~\ref{fig:ablation}-(a) shows the results of varying the location of a fixed-size capture region. The results indicate that \ourmethod consistently produces similar texture patterns, regardless of the location of the captured region.

\subsubsection{Effect of the capture scale.}
\begin{figure*}[t]
     \centerline{\includegraphics[width=0.98\linewidth]{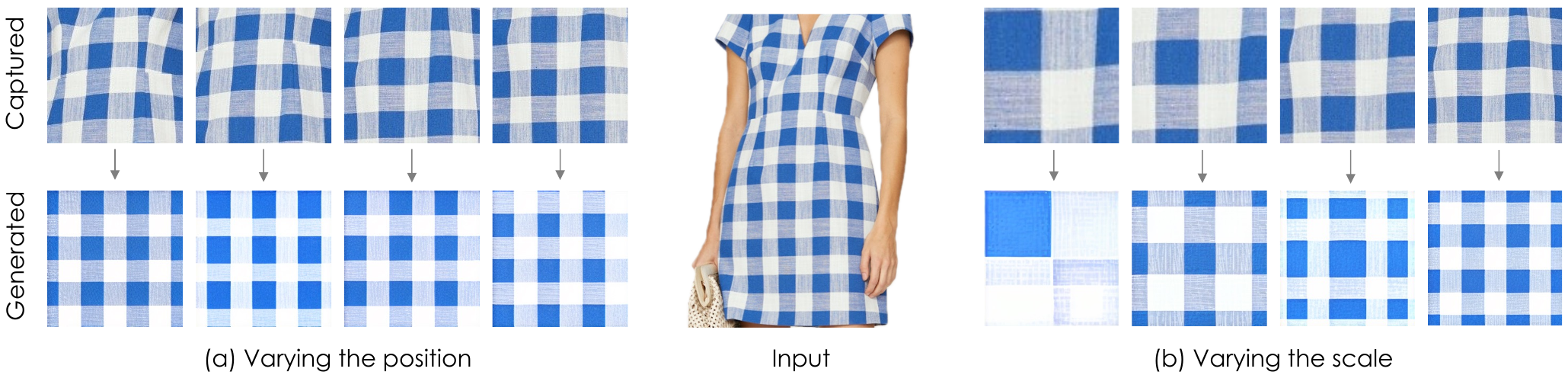}}
    \vspace{-2mm}
    \caption{\small Ablation study on varying the position and scale of the captured texture. Given an input clothing image, we evaluate (a) varying the position with a fixed capture size and (b) varying the scale for texture extraction. Our method successfully recovers the input texture despite variation in the location or resolution of the captured image. Since we care about distributions, none of the generated images are cherry- or lemon-pick.
    }
    \label{fig:ablation} 
    \vspace{3mm}
\end{figure*}

In Figure~\ref{fig:ablation}-(b), we further study the effect of the size of the captured region. By varying the scale of the captured region, \ourmethod recovers the texture pattern from the input patch, demonstrating robustness to changes in resolution. 

\input{table/tab_ablation_pseudo_brdf}

\subsubsection{Multi-material texture transfer.}
\begin{figure*}[h]
    \centering
     \includegraphics[width=0.98\linewidth]{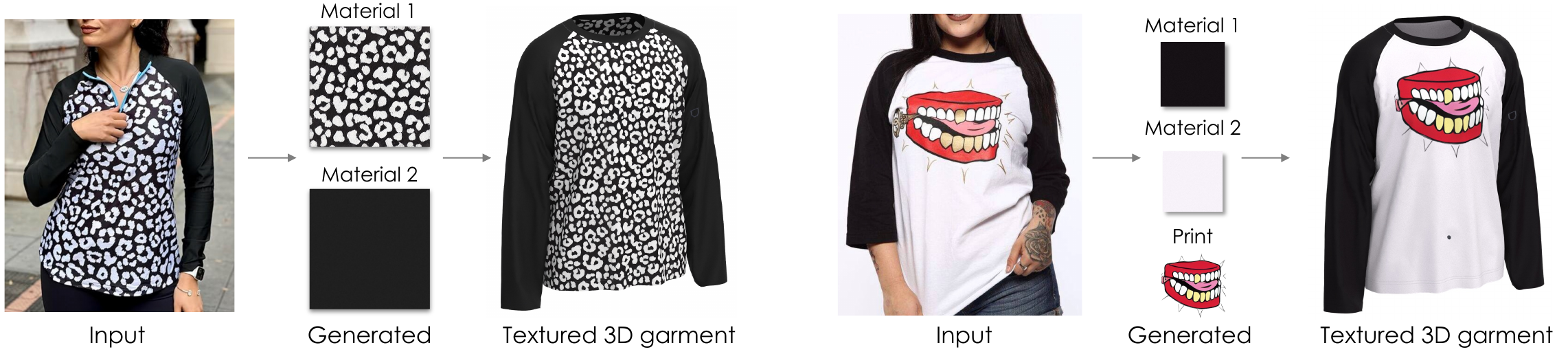}
    \vspace{-3mm}
    \caption{\small Multi-material textures transfer. Given a clothing image containing multiple texture patterns, materials, and prints, \ourmethod can transfer each distinct element to separate regions of the target 3D garment.
    }
    \label{fig:exp_multi} 
    \vspace{3mm}
\end{figure*}

Since \ourmethod works on patches, it can be applied to multi-material garments as well as evidenced in Figure~\ref{fig:exp_multi}. This suggests that \ourmethod can serve as a basic building block for multi-material garment texture transfer.

\subsubsection{Compatibility with AI-Generated Images.}
\begin{figure*}[t]
     \centerline{\includegraphics[width=0.95\linewidth]{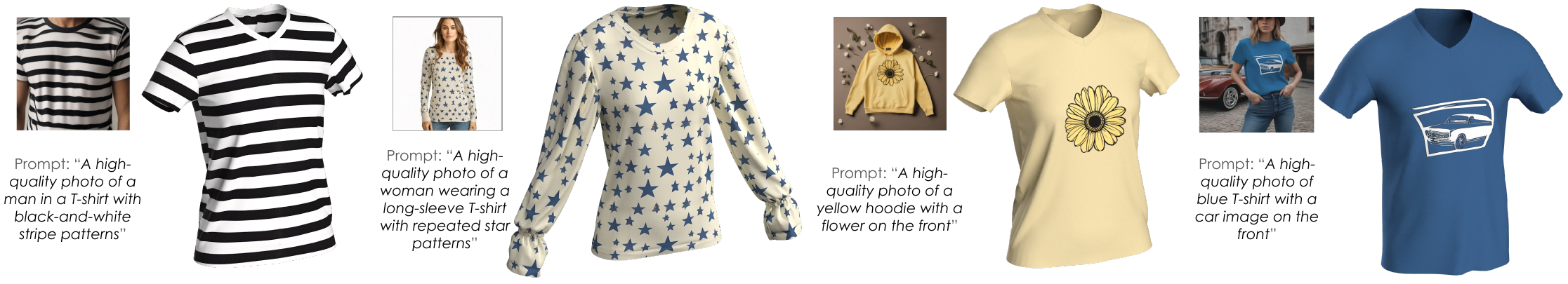}}
    \vspace{-3mm}
    \caption{\small Compatibility with generative apparel. \ourmethod can extract the textures from the output image of a text-to-image generative model and apply them to a target 3D garment of arbitrary shapes. We highlight that our method can handle imperfect textures, such as the broken black stripes in the first example. For each example, we show the input text prompt (bottom-left), the generated 2D image by Stable Diffusion XL (top-left), and the textured 3D garment (right) created by our \ourmethod method.
    }
    \label{fig:exp_genai} 
    \vspace{3mm}
\end{figure*}

We explore the possibility of enhancing \ourmethod with AI-generated images and demonstrate the results in Figure~\ref{fig:exp_genai}. In addition to real-life clothing, we use an advanced text-to-image model to create apparel images and the apply \ourmethod to transfer their textures to the target 3D garments. 
This opens up new creative possibilities for designers, allowing them to envision and materialize entirely novel textures and patterns through simple text descriptions.

\begin{figure*}[h]
    \centering
     \includegraphics[width=1\linewidth]{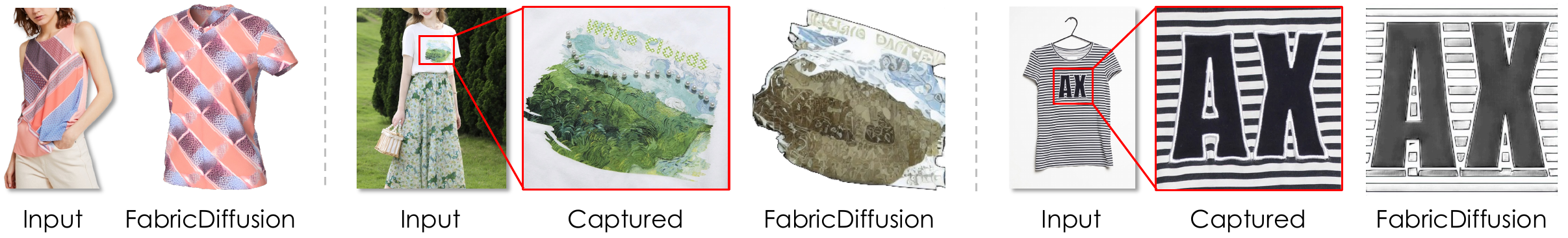}
    \vspace{-5mm}
    \caption{\small Limitations of \ourmethod. Our method may struggle to reconstruct specific inputs such as complex (e.g., non-repetitive) patterns (left), fine details in complex prints (middle), and prints over non uniform fabric (right).
    }
    \label{fig:limitation} 
\end{figure*}

%% file: table/tab_rendered_clothing.tex
\begin{table}[t]
\small
\tabcolsep 1.5pt 
\centering
\caption{\small Quantitative comparison on image-to-garment clothing texture transfer. Performances evaluated on synthetic testing data. Our method succeeds at faithfully extracting and transferring textures from images, whereas Material Palette~\cite{lopes2023material} exhibits significant artifacts, resulting in suboptimal performance, particularly on FID.}
\vspace{-3mm}
\begin{tabular}{lcccccc}
\toprule
 & FID $\downarrow$ & LPIPS$\downarrow$ & SSIM$\uparrow$ & MS-SSIM$\uparrow$  & DISTS$\downarrow$ & CLIP-s$\uparrow$  \\
\midrule
Material Palette & 34.39 & 0.20 & 0.75 & 0.73 & 0.28 & 0.94 \\
FabricDiffusion (ours) & \textbf{12.44} & \textbf{0.16} & \textbf{0.79} & \textbf{0.77} & \textbf{0.19} & \textbf{0.97} \\
\bottomrule
\end{tabular}\label{tab:sota_entire_clothing}
\end{table}

%% file: table/tab_pbr.tex
\begin{table}[t]
\small
\tabcolsep 2.5pt 
\centering
\caption{\small Quantitative comparison with state-of-the-art methods on PBR material extraction. Results are evaluated on the real PBR test examples. By fine-tuning MatFusion with additional fabric PBR training data, our method achieves superior performance across most material maps. Material Palette performs subpar, particularly in estimating the diffuse and roughness maps, due the differences in physical properties between fabric materials and general objects. Please see Table~\ref{tab:rendered_image} for quantitative evaluation on rendered images and Figure~\ref{fig:exp_pbr} for a qualitative comparison between \ourmethod and Material Palette.}
\vspace{-3mm}
\begin{tabular}{l|ccc|ccc}
\toprule
 &  \multicolumn{3}{c}{MSE$\downarrow$} & \multicolumn{3}{|c}{SSIM$\uparrow$} \\
 &  Diff. & Norm. & Rough. & Diff. & Norm. & Rough. \\
 \midrule
Material Palette  & \underline{0.0515} & 0.0136 & 0.1287 & 0.2213 &	0.3028 &	0.2920 \\
MatFusion  & 0.0896 & \underline{0.0127} & \underline{0.0806} & \underline{0.2190} & \textbf{0.3902} & \underline{0.4922} \\
FabricDiffusion (ours)  & \textbf{0.0287} & \textbf{0.0094} & \textbf{0.0559} & \textbf{0.3157} & \underline{0.3827} & \textbf{0.5039} \\
\bottomrule
\end{tabular}\label{tab:sota_pbr}
\end{table}

%% file: table/tab_render_metrics.tex
\begin{table}[t]
\small
\tabcolsep 4.5pt 
\centering
\caption{{\small Quantitative comparison on rendered materials. We adopt render-aware and perceptual metrics and compare the quality of rendered generated texture. \ourmethod outperforms other methods.}}
\vspace{-3mm}
\begin{tabular}{lcccccc}
\toprule
 & MSE$\downarrow$ & SSIM$\uparrow$ & DISTS$\downarrow$  & LPIPS$\downarrow$ & FLIP$\downarrow$  \\
\midrule
Material Palette & \underline{0.0531}	& 0.2838	& \underline{0.3388}	& \underline{0.4463} &	\underline{0.5812} \\
MatFusion & 0.1032	& \underline{0.3233}	& 0.3790	& 0.5697	& 0.7009 \\
FabricDiffusion (ours) & \textbf{0.0284}	& \textbf{0.4102}	 & \textbf{0.3035}	& \textbf{0.3836} & \textbf{0.4411} \\
\bottomrule
\end{tabular}\label{tab:rendered_image}
\end{table}

%% file: table/tab_ablation_pseudo_brdf.tex
\begin{table}[t]
\small
\tabcolsep 5.5pt 
\centering
\caption{{\small Ablation study on pseudo-BRDF data. We compare the performance of using combined versus only real-BRDF data. Combined data effectively improve the performance.}}
\vspace{-3mm}
\begin{tabular}{cc|cccc}
\toprule
  Real-BRDF & Pseudo-BRDF & FID$\downarrow$ & LPIPS$\downarrow$ & DISTS$\downarrow$ & CLIP-s$\uparrow$ \\
\midrule
 \checkmark &  & 19.17 & 0.19 & 0.26 & 0.96 \\
  \checkmark & \checkmark & \textbf{12.44} & \textbf{0.16} & \textbf{0.19} & \textbf{0.97} \\
\bottomrule
\end{tabular}\label{tab:pseudo-BRDF}
\end{table}

%% file: section/disc.tex
\section{Discussion, Limitation, and Conclusion}

In this paper, we introduce \ourmethod, a new method for transferring fabric textures and prints from a single real-world clothing image onto 3D garments with arbitrary shapes. 
Our method, trained entirely using {synthetic} rendered images, is able to generate undistorted texture and prints from in-the-wild clothing images.
While our method demonstrates strong generalization abilities with real photos and diverse texture patterns, it faces challenges with certain inputs, as shown in Figure~\ref{fig:limitation}. Specifically, \ourmethod may produce errors when reconstructing non-repetitive patterns and struggles to accurately capture fine details in complex prints or logos, especially since our focus is on prints with uniform backgrounds, moderate complexity, and moderate distortion. 
In the future, we plan to address these challenges by enhancing texture transfer for more complex scenarios and improving performance on difficult fabric categories, such as leather. Additionally, we plan to expand our method to handle a broader range of material maps, including transmittance, to further extend its applicability.

%% file: section/app.tex
\section{Key Advantages of \ourmethod}
\label{sec:appendixA}

\paragraph{Normalized texture representation.}
Unlike existing image-to-3D texture transfer methods, \ourmethod generates normalized textures that can be used in the 2D UV space. We highlight two outputs: (1) High-quality, distortion-free, and tileable texture maps from a non-rigid garment surface. (2) Seamless integration with SVBRDF material estimation pipelines, which usually build upon the first output --- standard close-up views of the materials as input.

\paragraph{Sim-to-real generalizability.}
The conditional diffusion model, trained entirely using \emph{synthetic} rendering images, proves highly effective in generating normalized texture maps from \emph{real-world} images. We attribute this success to: (1) Our model bridging the domain gap between real and rendered textures by conditioning on the real input texture. (2) Synthetic data offering controllable supervision and diverse geometric, illumination, and occlusion variations. 

\paragraph{Data and computational efficiency.}
During training, our method of creating pseudo-BRDF material is effective in scaling up the training examples. During inference, our model performs feed-forward sampling from Gaussian noise, which takes approximately less than 5 seconds on a single NVIDIA A6000 GPU. In contrast, existing texture transfer methods often rely on costly per-example optimization.

\section{Details on Dataset Construction}
\label{sec:appendixB}

\paragraph{Fabric BRDF and textile dataset.}
To curate textures and their BRDF materials, we use several public libraries (AmbientCG\footnote{https://ambientcg.com/}, ShareTextures\footnote{https://www.sharetextures.com/},  3D Textures\footnote{https://3dtextures.me/}) under the CC0 license and supplement them with additional assets purchased from artists. The real BRDF dataset we collected comprises 3.8k assets, encompassing a broad spectrum of fabric materials. The pseudo-BRDF dataset contain 100k fabric textures with only RGB color images. We reserved 200 materials from the real BRDF dataset for testing our BRDF generator, and 800 materials from the pseudo BRDF dataset (combined with the previous 200 materials) for testing the texture flattening module. 

Our textile images are collected from online sources including Openverse\footnote{https://openverse.org/}, PublicDomainPictures\footnote{https://publicdomainpictures.net/en/}, and ARTX\footnote{https://architextures.org/} under CC0 or royalty-free license.

\paragraph{3D garment mesh dataset.}
We collect 22 raw 3D garment meshes for training and 5 garment meshes for testing. That is, during the testing with synthetic data, the model has not seen the geometry from the 5 testing meshes. With the method described in Section 3.2 of the main paper, we construct approximately 220k flat and warped texture pairs for training and 5k pairs for testing.

\paragraph{Logos and prints dataset.}
We collect a dataset of 7k prints and logos in PNG format with CC0 license. Their corresponding pseudo-BRDF materials are generated by assigning a uniform roughness value sampled from $\mathcal{U}(0.4,0.7)$, a uniform metallic value sampled from $\mathcal U(0,0.3)$, and a default flat normal map. In cases where a print was uniformly black, we converted it to white if the background texture was also dark. By compositing the logo prints onto the 3D garments, we obtain a total of 82k warped print images, following the method outlined in Section~~\ref{s_method_data} of the main paper.

\section{Additional Details of Our Method}
\label{sec:appendixC}

\subsection{Details on physics-based rendering}
During rendering, each image pixel value at a specific viewing direction can be computed using the following reflectance equation:
\begin{equation} \label{eq:rendering_equation}
L(p,\omega_o)=\int_{\Omega} f_r(p,\omega_{i},\omega_o) L_{i}(p,\omega_{i}) (\omega_{i} \cdot n_p) \mathrm{d}\omega_{i},
\end{equation}
where $L$ is the rendered pixel color along the direction $\omega_o$ from the surface point $p$, $\Omega = \{ \omega_i: \omega_i \cdot n_p \geq 0 \}$ denotes a hemisphere with the incident direction $\omega_i$ and surface normal $n_p$ at point $p$, $L_i$ is the incident light that is represented by the environment map, and $f_r$ is known as the BRDF that scales or weighs the incoming radiance based the material parameters $(k_d, k_n, k_r, k_m)$ of the garment surface. By aggregating the rendered pixel colors along the direction $\omega_o$ (i.e., camera pose), we are able to obtain the rendered image of the input patch (image $x$ in Equation~~\ref{eq:statement} of the main paper). 

\subsection{Classifier-free guidance for conditional image generation}
We leverage Classifier-Free Guidance (CFG)~\cite{ho2022classifier} during the training for trading off the quality and diversity of samples generated by our \ourmethod model. The implementation of CFG involves jointly training the diffusion model for conditional and unconditional denoising, and combining the two score estimates (the $\ell_2$ loss of the noise term in Equation (3) of the main paper) at inference time. Training for unconditional denoising is done by simply setting the conditioning to a fixed null value $\mathcal{E}(x)~{=}~\varnothing$ at some frequency during training. At inference time, with a guidance scale $s\ge1$, the modified score estimate $\tilde{e_{\theta}}(x_t, \mathcal{E}(x))$ is extrapolated in the direction toward the conditional $e_{\theta}(x_t, \mathcal{E}(x))$ and away from the unconditional $e_{\theta}(x_t, \varnothing)$:
\begin{equation}
    \tilde{e_{\theta}}(x_t, \mathcal{E}(x)) = e_{\theta}(x_t, \varnothing) + s \cdot (e_{\theta}(x_t, \mathcal{E}(x)) - e_{\theta}(x_t, \varnothing)).
    \label{eq:cfg}
\end{equation}
CFG enhances the visual quality of generated texture maps and ensures that the sampled images more accurately correspond to the input texture in terms of color, pattern, and scale.

\subsection{Strategy for determining tiling scales}
After extracting PBR material maps from an image exemplar, we tile them in the garment UV space for realistic rendering. The key question is {how to determine the scale for tiling?} We investigate two specific strategies: (1) Proportion-aware tiling. We use image segmentation to calculate the proportion of the captured region relative to the segmented clothing, maintaining the same ratio when tiling the generated texture onto the sewing pattern. (2) User-guided tiling. We emphasize that an end-to-end automatic tilling method may not be optimal, as user involvement is often necessary to resolve ambiguities and provide flexibility in fashion industries. 

\subsection{Implementation details}
We use pre-trained Stable Diffusion v1.5 as the backbone of the normalized texture map generation and finetune it on our texture and print datasets, respectively. Both the input and output scales are set as 256$\times$256px. We use a batch size of 512 and a learning rate of $5\times10^{-5}$. It takes roughly 2 days (20k iterations) to train on four NVIDIA A6000 GPUs. For PBR materials estimation, we fine-tuned the pre-trained MatFusion model for roughly 1 hour with our 3.8k BRDF materials training data.

\section{Additional Results}\label{s_add_exp}

\input{table/tab_synthetic_clothing_to_texture}

\begin{figure}[t]
     \centerline{\includegraphics[width=1\linewidth]{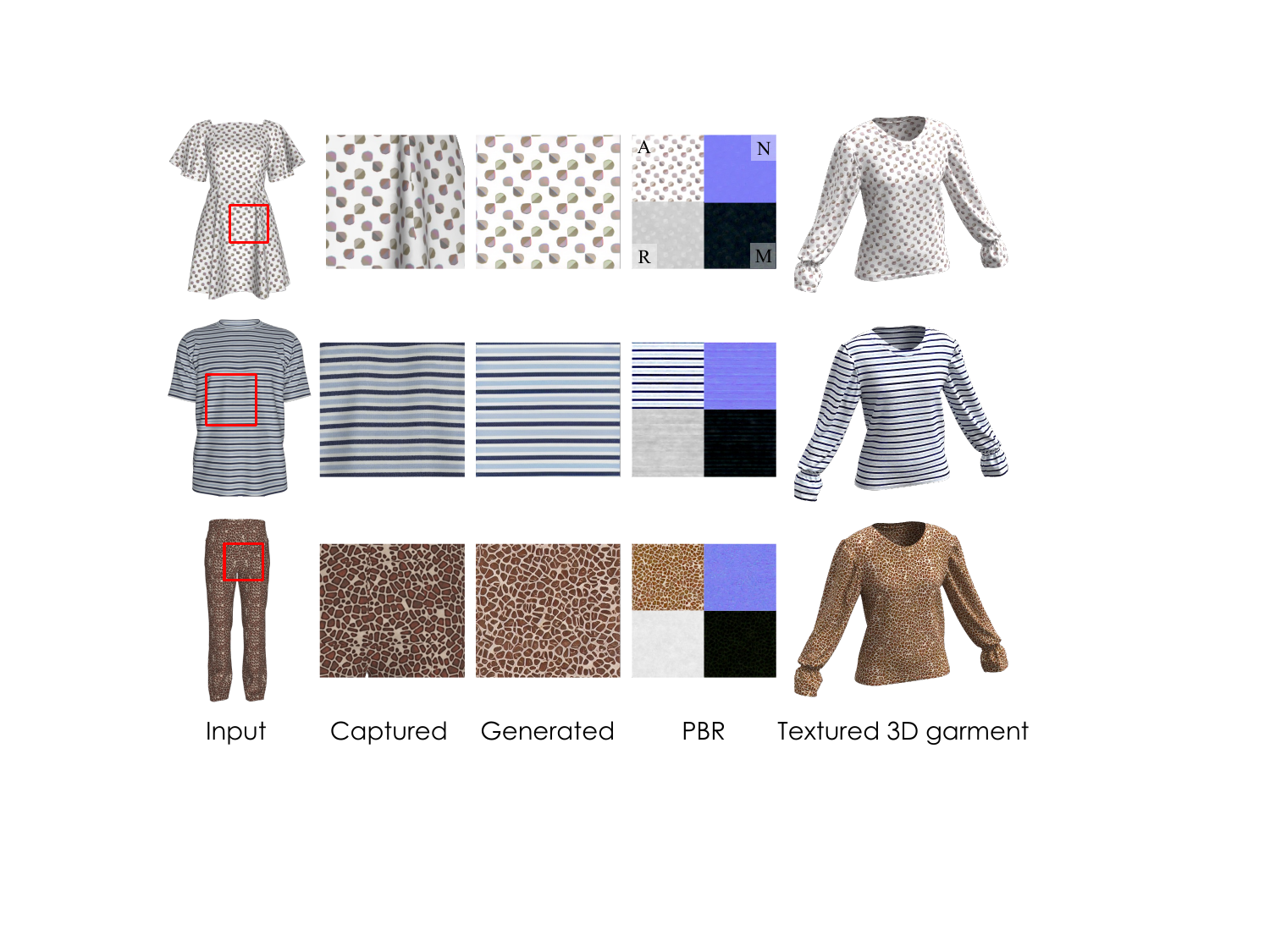}}
    \caption{\small Texture transfer on synthetic data. Given the input image of the 3D garment and a captured patch, our method generates a normalized texture map that is flat and tileable, along with the corresponding PBR materials. The PBR materials maps can be applied to the target 3D garment with different geometry for reliable rendering. Our model is capable of removing shadows (1st row), disentangling distortions (1st \& 2nd row), and capturing physical properties (3rd row) from the input fabric texture. Note that, both the input 3D garment meshes and textures in this figure were not used for model training. See Table~\ref{tab:sota_entire_clothing} of the main paper for qualitative results.
    }
    \label{fig:exp_synthetic} 
\end{figure}

\subsection{Additional results on textures extraction}
Generating a normalized texture image plays a crucial intermediate step to ensure reliable texture transfer. Figure~\ref{fig:exp_pbr} (in the main paper) shows some cases of the generated normalized textures. In Table~\ref{tab:sota_clothing_texture}, we provide a quantitative analysis using synthetic data, for which we have ground-truth textures, and compare our method with state-of-the-are methods. As we observe, our method consistently outperforms Material Palette~\cite{lopes2023material} across various evaluation metrics. As discussed in Section~\ref{s_related} and Section~\ref{s_exp_main} of the main paper, personalization-based methods struggle at capturing fine-grained texture details, or disentangling the effects of distortion.

\subsection{Texture transfer on synthetic data}
We also validate our method using synthetic data and show the qualitative results in Figure~\ref{fig:exp_synthetic}. We test on textured garments with ground-truth BRDF materials, enabling controlled evaluation of geometric distortions and illumination variations. Our method reliably generates normalized textures and PBR materials. As our focus is on clothing fabrics with minimal metallic properties, we omit metallic map results for simplicity in the following experiments. Quantitative results are shown in Table~\ref{tab:sota_entire_clothing} of the main paper.

%% file: table/tab_synthetic_clothing_to_texture.tex
\begin{table}[t]
\small
\tabcolsep 3.5pt 
\centering
\caption{\small Quantitative comparison on texture images extraction from 3D garments. Results are evaluated on synthetic testing data. The ground-truths are normalized texture images that are flat and with a unified lighting condition. Our method outperforms Material Palette~\cite{lopes2023material} across different evaluation metrics.}
\begin{tabular}{lccccc}
\toprule
 & LPIPS$\downarrow$ & SSIM$\uparrow$ & MS-SSIM$\uparrow$ & DIST$\downarrow$ & CLIP-s$\uparrow$ \\
\midrule
Material Palette & 0.66 & 0.27 & 0.31 & 0.45 & 0.89 \\
FabricDiffusion (ours) & \textbf{0.53} & \textbf{0.32} & \textbf{0.32} & \textbf{0.32} & \textbf{0.91} \\
\bottomrule
\end{tabular}\label{tab:sota_clothing_texture}
\end{table}